# Algorithms for Bayesian network modeling and reliability inference of complex multistate systems: Part II – Dependent systems

Xiaohu Zheng[a,b], Wen Yao[b*], Xiaoqian Chen[b]

[a] *College of Aerospace Science and Engineering, National University of Defense Technology, No. 109 Deya Road, Kaifu District, Changsha, Hunan Province, China 410073*

[b] *Defense Innovation Institute, Chinese Academy of Military Science, No.53, East Main Street, Fengtai District, Beijing, China 100071*

* Corresponding Author: wendy0782@126.com

**Abstract**

In using the Bayesian network (BN) to construct the complex multistate system's reliability model as described in Part I, the memory storage requirements of the node probability table (NPT) will exceed the random access memory (RAM) of the computer. However, the proposed inference algorithm of Part I is not suitable for the dependent system. This Part II proposes a novel method for BN reliability modeling and analysis to apply the compression idea to the complex multistate dependent system. In this Part II, the dependent nodes and their parent nodes are equivalent to a block, based on which the multistate joint probability inference algorithm is proposed to calculate the joint probability distribution of a block's all nodes. Then, based on the proposed multistate compression algorithm of Part I, the dependent multistate inference algorithm is proposed for the complex multistate dependent system. The use and accuracy of the proposed algorithms are demonstrated in case 1. Finally, the proposed algorithms are applied to the reliability modeling and analysis of the satellite attitude control system. The results show that both Part I and Part II's proposed algorithms make the reliability modeling and analysis of the complex multistate system feasible.

**Nomenclature**

| | | | |
|---|---|---|---|
| $C_L^S$ | the $L$th parent node of node $S$, where $L=1,2,\cdots,\gamma$ ($\gamma \geq 2$) | $\alpha$ | the number of blocks in BN |
| | | $m_{j_H}$ | the number of the root nodes in the |
| | $j_H$ th block, $j_H = 1, 2, \cdots, \alpha$ | $c\lambda_i$ | the compressed $\lambda_i$ |
| $n_{\bar{H}}$ | the number of independent nodes in BN | $c\lambda_i^j$ | the $j$ th row of $c\lambda_i$ |
| $n_{j_H}$ | the number of the child nodes in the $j_H$ th block | $d_{r_i}$ | the run accompanying dictionary of $c\lambda_i$ |
| $H_{j_H}^{h_{j_H}}$ | the $h_{j_H}$ th root node in the $j_H$ th block, $h_{j_H} = 1, 2, \cdots, m_{j_H}$ | $d_{p_i}$ | the phrase accompanying dictionary of $c\lambda_i$ |
| $C_{j_H}^{c_{j_H}}$ | the $c_{j_H}$ th child node in the $j_H$ th block, $c_{j_H} = 1, 2, \cdots, n_{j_H}$ | $q_i^j$ | the row index of a run in $d_{r_i}$ |
| | | $n_{r_i}^j$ | the number of repeated instances of the run in $\lambda_i$ |
| $C_{i_{\bar{H}}}$ | the $i_{\bar{H}}$ th independent node, $i_{\bar{H}} = 1, 2, \cdots, n_{\bar{H}}$ | $r_i^j$ | the numerical value consisted the run in the $q_i^j$ th row of $d_{r_i}$ |
| $\pi(C_{j_H}^{c_{j_H}})$ | all parent nodes of $C_{j_H}^{c_{j_H}}$ | $L_{r_i}^j$ | the number of $r_i^j$ |
| $N_{j_H}^{h_{j_H}}$ | the state number of the root node $H_{j_H}^{h_{j_H}}$ | $p_i^j$ | the row index of a phrase in $d_{p_i}$ |
| $k_b$ | the row index of the joint probability table, $k_b = 1, 2, \cdots, T_{j_H}$ | $n_{p_i}^j$ | the number of repeated instances of the phrase in $\lambda_i$ |
| $s_{i_b}^{k_b}$ | for $\{C_{j_H}^1, \cdots, C_{j_H}^{n_{j_H}}, H_{j_H}^1, \cdots, H_{j_H}^{m_{j_H}}\}$, the state of the $i_b$ th node in the $k_b$ th row of the joint probability table, $i_b = 1, 2, \cdots, l$ ($l = n_{j_H} + m_{j_H}$) | $v_{1_i}^j$ | the first numerical value consisted the phrase in the $p_i^j$ th row of $d_{p_i}$ |
| | | $v_{2_i}^j$ | the second numerical value consisted the phrase in the $p_i^j$ th row of $d_{p_i}$ |
| | | $L_{p_i}^j$ | the number of $v_{1_i}^j$ and $v_{2_i}^j$ |
| $M_{j_H}^{c_{j_H}}$ | the state number of the child node $C_{j_H}^{c_{j_H}}$ | $S_{i+1}^j$ | the start row number of the run or phrase in $\lambda_{i+1}$ defined by row $j$ of $c\lambda_{i+1}^j$ |
| $T_{j_H}$ | the state combination number of all the nodes in the $j_H$ th block | $S_{i+1}^{all}$ | the start row number set of all runs and phrases in $\lambda_{i+1}$ |
| $T_{j_H}^C$ | the state combination number of all child nodes in the $j_H$ th block | $RP^{i+1}$ | the set of $RP_{i+1}^j$, i.e. $RP^{i+1} = \{RP_{i+1}^1, RP_{i+1}^2, \cdots, RP_{i+1}^j, \cdots, RP_{i+1}^{m_{i+1}}\}$ |
| $\Pr_{j_H}$ | the joint probability of all child nodes in the $j_H$ th block | | |
| $p_k$ | the the $k$ th row's probability of the joint probability table | $RP_{i+1}^j(1, i_{RP})$ | the $i_{RP}$ th element of $RP_{i+1}^j$ |
| $i$ | the subscript index | $J_{i+1}^j$ | the set that includes all positions of all elements of $RP_{i+1}^j$ in the set $S_{i+1}^{all}$ |
| $q_i^{\max 2}$ | the max row index in the run accompanying dictionary $d_{r_i}$ | $L_{RP}$ | the element number of $RP_{i+1}^j$ |
| | | $J_{i+1}^j(1, i_{RP})$ | the $i_{RP}$ th element of $J_{i+1}^j$, $i_{RP} = 1, 2, \cdots, L_{RP}$ |
| $\lambda_i$ | the joint probability after eliminating the $i$ th node of the joint probability table | $R_{i+1}^j$ | the remainder that is obtained after |
| | finishing the construction of $c\lambda_i^j$ based on $c\lambda_{i+1}^j$ | $c\lambda_i^{new}$ | the new compressed intermediate factor after eliminating the $i$ th node |
| $R^{all}$ | the set of $R_{i+1}^j$ | $d_{r_i}^{new}$ | the run accompanying dictionary of $c\lambda_i^{new}$ |
| $R_{J_{i+1}^j(1,i_{RP})}^{all}$ | the $J_{i+1}^j(1,i_{RP})$ th element of $R^{all}$ | $d_{p_i}^{new}$ | the phrase accompanying dictionary of $c\lambda_i^{new}$ |
| $I$ | the position of $S_{i+1}^j$ in $S_{i+1}^{all}$ | | |
| $R_{I-1}^{all}$ | the $(I-1)$ th element of $R^{all}$ | $j_u$ | the row index of $c\lambda_{i+1}$, $j_u = 1,2,\cdots,m_{i+1}$ |
| $E$ | the node evidence set | $c\lambda_i^{new-j_u}$ | the $j_u$ th row of $c\lambda_i^{new}$ |
| $Q$ | the query node set | | |
| $RP_{i+1}^{update}$ | the updated $RP^{i+1}$ | $d_{r_i}^{new-j_u}$ | the run in the $j_u$ th row of $c\lambda_i^{new}$ |
| $S_{i+1}^{update}$ | the updated $S_{i+1}^{all}$ | $d_{p_i}^{new-j_u}$ | the phrase in the $j_u$ th row of $c\lambda_i^{new}$ |

## 1. Introduction

As described in Part I [1], the memory storage requirements of the node probability table (NPT) [2, 3] will exceed the computer's random access memory (RAM) when the components reach a certain amount. Therefore, Part I proposed a multistate compression algorithm to compress the NPT. Based on Part I's studies, this Part II studies the application of compression idea in the Bayesian network (BN) modeling and reliability inference for the complex multistate dependent system.

With the increasing complexity of engineering systems, some components of the complex system are dependent. By modeling the probabilistic dependencies between components, O'Connor and Mosleh [4] proposed a general dependency model for system reliability analysis based on BN. Considering common cause failure factors, Mi et al. [5-8] proposed system reliability analysis methods based on BN. For the complex system with dependent outages, Wojdowski and Anders [9] proposed a BN method for the substation reliability assessment. For the dependent system that risk factors interact with each other, Pan et al. [10] developed a hybrid Copula-Bayesian approach for risk modeling and evaluation based on BN. However, the above methods for dependent system can not solve the NPT's memory storage requirements problem.

To solve the dependent system's memory storage problem, Tien et al. [2] discussed the treatment of complex systems with dependent components. However, Tien's algorithm for dependent components is only suitable for the complex binary system. Besides, for the methods of both Zheng et al. [3] and Tong et al. [11], as discussed in Part I [1], in addition to being restricted by system state configurations, their methods cannot be suitable for the multistate system with dependent components. Therefore, based on the proposed two algorithms of Part I, this Part II proposes a novel dependent multistate inference algorithm for the reliability analysis of the complex multistate dependent system. Not limited by the system structure and state configurations, the proposed dependent multistate inference algorithm can be applied to any complex multistate system.

The rest of this paper (Part II) is organized as follows. In section 2, a block is defined for processing the dependent nodes of BN, and two properties of the block are discussed at the same time. Section 3 includes four subsections: section 3.1 expounds which rules and algorithms of Part I are used in this Part II, and the rules for constructing dependent intermediate factors and the multistate joint probability inference algorithm are proposed respectively in section 3.2 and 3.3, and the dependent multistate inference algorithm is proposed for performing the inference of system BN reliability model in section 3.4. In section 4, the proposed algorithms are validated by the reliability analysis of a pitch axis subsystem and a satellite attitude control system, respectively. Conclusions and future researches are discussed in section 5.

## 2 Block in BN

*2.1 Block*

In BN, if a node has no parent nodes, this node is a root node like node A, node B, and node C in Fig. 1. If a node has no child nodes, this node is a leaf node like the node E in Fig. 1.

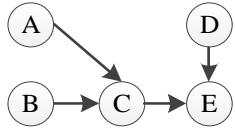

Fig. 1 BN for explaining the definitions of the root node and leaf node

Given the following four conditions: (1) BN only has one leaf node, denoted as $S$. (2) Suppose that node $S$ has $\gamma$ ($\gamma \geq 2$) parent nodes, marked as $\{C_1^S, C_2^S, \cdots, C_L^S, \cdots, C_\gamma^S\}$, where $L = 1, 2, \cdots, \gamma$. If any node $C_L^S$ in $\{C_1^S, C_2^S, \cdots, C_L^S, \cdots, C_\gamma^S\}$ has parent nodes, these parent nodes of the node $C_L^S$ must be root nodes. (3) There is no directed arc between any two nodes in $\{C_1^S, C_2^S, \cdots, C_L^S, \cdots, C_\gamma^S\}$. (4) There is no directed arc between any two root nodes in BN.

For the BN that meets the above four conditions, suppose that $\beta$ ($\beta \geq 2$) nodes, denoted as $\{C_1^H, C_2^H, \cdots, C_l^H, \cdots, C_\beta^H\}$, meet the following two conditions:

**Condition 1**: Node $C_l^H$ ($l = 1, 2, \cdots, \beta$) has one parent node at least.

**Condition 2**: $\exists C \in \{C_1^H, C_2^H, \cdots, C_{l-1}^H, C_{l+1}^H, \cdots, C_\beta^H\}$, there is at least one common parent node between $C_l^H$ and $C$.

Therefore, $\beta$ nodes and their parent nodes form a "**block**". In the BN that meets the above four conditions, a block has the following two properties.

**Property 1**: There are no common nodes between any two blocks, i.e., any two blocks are independent of each other.

**Property 2**: If the parent node $C_L^S$ of the leaf node $S$ is a root node, $C_L^S$ and any one block are independent to each other.

Suppose that a BN has $\alpha$ blocks and $n_{\bar{H}}$ independent nodes, as shown in Fig. 2. For the $j_H$th block, it has $m_{j_H}$ root nodes $\{H_{j_H}^1, H_{j_H}^2, \cdots, H_{j_H}^{m_{j_H}}\}$ and $n_{j_H}$ child nodes $\{C_{j_H}^1, C_{j_H}^2, \cdots, C_{j_H}^{n_{j_H}}\}$, where $j_H = 1, 2, \cdots, \alpha$. In the $j_H$th block, $H_{j_H}^{h_{j_H}}$ means the $h_{j_H}$th root node and $C_{j_H}^{c_{j_H}}$ means the $c_{j_H}$th child node, where $h_{j_H} = 1, 2, \cdots, m_{j_H}$ and $c_{j_H} = 1, 2, \cdots, n_{j_H}$. $C_{i_{\bar{H}}}$ is the $i_{\bar{H}}$th ($i_{\bar{H}} = 1, 2, \cdots, n_{\bar{H}}$) independent node. The joint probability of all nodes in the $j_H$th block can be calculated by the following equation (1), where $\pi(C_{j_H}^{c_{j_H}})$ denotes all parent nodes of $C_{j_H}^{c_{j_H}}$.

$$\Pr(C_{j_H}^1, \cdots, C_{j_H}^{n_{j_H}}, H_{j_H}^1, \cdots, H_{j_H}^{m_{j_H}}) = \prod_{c_{j_H}=1}^{n_{j_H}} \Pr(C_{j_H}^{c_{j_H}} \mid \pi(C_{j_H}^{c_{j_H}})) \prod_{h_{j_H}=1}^{m_{j_H}} \Pr(H_{j_H}^{h_{j_H}}) \quad (1)$$

Therefore, the joint probability of all child nodes in the $j_H$th block can be calculated by the following equation (2), where $N_{j_H}^{h_{j_H}}$ means the state number of the root node $H_{j_H}^{h_{j_H}}$.

$$\Pr(C_{j_H}^1, \cdots, C_{j_H}^{n_{j_H}}) = \sum_{h_{j_H}=1}^{m_{j_H}} \sum_{H_{j_H}^{h_{j_H}}=1}^{N_{j_H}^{h_{j_H}}} \Pr(C_{j_H}^1, \cdots, C_{j_H}^{n_{j_H}}, H_{j_H}^1, \cdots, H_{j_H}^{m_{j_H}}) \quad (2)$$

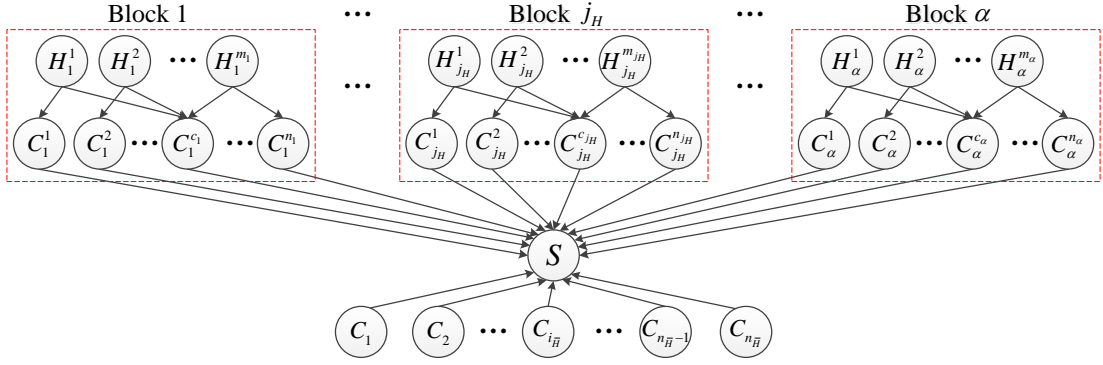

Fig. 2 A system BN has $\alpha$ blocks and $n_{\bar{H}}$ independent nodes

*2.2 Equivalent processing of block*

Suppose that the child node $C_{j_H}^{c_{j_H}}$ has $M_{j_H}^{c_{j_H}}$ states. Therefore, the state combination number $T_{j_H}$ of all the nodes in the $j_H$ th block is

$$T_{j_H} = \prod_{h_{j_H}=1}^{m_{j_H}} N_{j_H}^{h_{j_H}} \prod_{c_{j_H}=1}^{n_{j_H}} M_{j_H}^{c_{j_H}} . \qquad (3)$$

For the child nodes $\{C_{j_H}^1, C_{j_H}^2, \cdots, C_{j_H}^{c_{j_H}}, \cdots, C_{j_H}^{n_{j_H}}\}$ in the $j_H$ th block, the state combination number $T_{j_H}^C$ of all child nodes is calculated by

$$T_{j_H}^C = \prod_{c_{j_H}=1}^{n_{j_H}} M_{j_H}^{c_{j_H}} . \qquad (4)$$

Given $n = n_{j_H}$, $i = c_{j_H}$ and $N_i = M_{j_H}^{c_{j_H}}$, the state of each node in the $k$ th row of the joint probability table of child nodes $\{C_{j_H}^1, C_{j_H}^2, \cdots, C_{j_H}^{c_{j_H}}, \cdots, C_{j_H}^{n_{j_H}}\}$ can be calculated by the rules in Part I [1] (The rules are shown in equation (4) and (6) of Part I). Besides, the joint probability of all child nodes in the $j_H$ th block is $\Pr_{j_H}$, i.e.

$$\Pr_{j_H} = \Pr(C_{j_H}^1, C_{j_H}^2, \cdots, C_{j_H}^{n_{j_H}}) . \qquad (5)$$

Therefore, according to the state combination in the $k$ th row, the joint probability $p_k$ corresponding to the row $k$ can be calculated by equation (2) and

$$\sum_{k=1}^{T_{j_H}^C} p_k = 1 . \qquad (6)$$

If the $j_H$ th block is equivalent to a node $C'_{j_H}$, each state combination of child nodes

$\{C_{j_H}^1, C_{j_H}^2, \cdots, C_{j_H}^{c_{j_H}}, \cdots, C_{j_H}^{n_{j_H}}\}$ is equivalent to a state of the node $C'_{j_H}$. Thus, the $j_H$ th block can be equivalent to a multistate node $C'_{j_H}$ with $T_{j_H}^C$ states. For the BN in Fig. 2, it can be equivalent to the BN in Fig. 3.

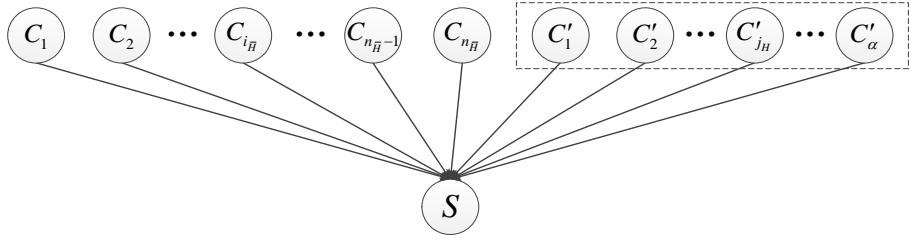

Fig. 3 The equivalent BN

According to **Property 1**, any two nodes of $\{C'_1, C'_2, \cdots, C'_\alpha\}$ are independent of each other. According to **Property 2**, $\forall C' \in \{C'_1, C'_2, \cdots, C'_\alpha\}$ and $\forall C_{i_{\bar{H}}} \in \{C_1, C_2, \cdots, C_{n_{\bar{H}}}\}$, $C'_{j_H}$ and $C_{i_{\bar{H}}}$ are independent to each other.

## 3. Proposed multistate inference algorithms

*3.1 Bases: rules and algorithms in Part I*

As described in Part I [1], the parent nodes' state combination in NPT's each row is calculated by the rules as shown in equations (4) and (6) of Part I. In addition to NPT, the nodes' state combination of the joint probability table's each row is also calculated by the above rules. Besides, both the NPT and the joint probability table are compressed by the **Multistate Compression Algorithm** of Part I. Based on the compressed NPT, the **Dependent Multistate Inference Algorithm** is proposed as shown in section 3.4. In particular, the independent node inference of BN is performed by the **Independent Multistate Inference Algorithm** of Part I. Therefore, the proposed **Dependent Multistate Inference Algorithm** of this Part II is based on the **Multistate Compression Algorithm** and the **Independent Multistate Inference Algorithm** in Part I.

*3.2 Rules for constructing dependent intermediate factors*

For the $j_H$ th block, denote that $k_b$ is the row index of the joint probability table of $\{C_{j_H}^1, \cdots, C_{j_H}^{n_{j_H}}, H_{j_H}^1, \cdots, H_{j_H}^{m_{j_H}}\}$ and $k_b = 1, 2, \cdots, T_{j_H}$. For the $k_b$ th row, the state of each node in $\{C_{j_H}^1, \cdots, C_{j_H}^{n_{j_H}}, H_{j_H}^1, \cdots, H_{j_H}^{m_{j_H}}\}$, denoted as $s_{i_b}^{k_b}$, can be calculated by the rules in Part I [1] (The rules are shown in equation (4) and (6) of Part I), where $i_b = 1, 2, \cdots, l$ and $l = n_{j_H} + m_{j_H}$. Thus, according to the state combination in the $k_b$ th row, the value of $\Pr(C_{j_H}^{c_{j_H}} | \pi(C_{j_H}^{c_{j_H}}))$ can be obtained by the relationship between the node $C_{j_H}^{c_{j_H}}$ and its parent nodes $\pi(C_{j_H}^{c_{j_H}})$. Finally, the joint probability value in the $k_b$ th row of the joint probability table of $\{C_{j_H}^1, \cdots, C_{j_H}^{n_{j_H}}, H_{j_H}^1, \cdots, H_{j_H}^{m_{j_H}}\}$ can be calculated by equation (1). The above process is equivalent to the function $\Psi\{C_{j_H}^{c_{j_H}}, \Pr(C_{j_H}^{c_{j_H}} | \pi(C_{j_H}^{c_{j_H}}))\}$. Therefore, given the $k_b$ th row's state combination of all nodes $\{C_{j_H}^1, \cdots, C_{j_H}^{n_{j_H}}, H_{j_H}^1, \cdots, H_{j_H}^{m_{j_H}}\}$, the joint probability $\Pr^{k_b}(C_{j_H}^1, \cdots, C_{j_H}^{n_{j_H}}, H_{j_H}^1, \cdots, H_{j_H}^{m_{j_H}})$ can be calculated by

$$\Pr^{k_b}(C_{j_H}^1, \cdots, C_{j_H}^{n_{j_H}}, H_{j_H}^1, \cdots, H_{j_H}^{m_{j_H}}) = \Psi\{C_{j_H}^{c_{j_H}}, \Pr(C_{j_H}^{c_{j_H}} | \pi(C_{j_H}^{c_{j_H}}))\}. \tag{7}$$

The joint probability $\Pr_{j_H}$ as shown in equation (5) is calculated by eliminating the root nodes $\{H_{j_H}^1, H_{j_H}^2, \cdots, H_{j_H}^{m_{j_H}}\}$ one by one, i.e.

$$\begin{aligned}
\Pr_{j_H} &= \sum_{H_{j_H}^1=1}^{N_{j_H}^1} \cdots \sum_{H_{j_H}^{m_{j_H}}=1}^{N_{j_H}^{m_{j_H}}} \Pr(C_{j_H}^1, \cdots, C_{j_H}^{n_{j_H}}, H_{j_H}^1, \cdots, H_{j_H}^{m_{j_H}}) &&\to \lambda_{m_{j_H}} \\
&\cdots &&\vdots \\
&= \sum_{H_{j_H}^1=1}^{N_{j_H}^1} \cdots \sum_{H_{j_H}^i=1}^{N_{j_H}^i} \Pr(C_{j_H}^1, \cdots, C_{j_H}^{n_{j_H}}, H_{j_H}^1, \cdots, H_{j_H}^i) &&\to \lambda_i \\
&\cdots &&\vdots \\
&= \sum_{H_{j_H}^1=1}^{N_{j_H}^1} \Pr(C_{j_H}^1, \cdots, C_{j_H}^{n_{j_H}}, H_{j_H}^1) &&\to \lambda_1 \\
&= \Pr(C_{j_H}^1, \cdots, C_{j_H}^{n_{j_H}}) &&\to \lambda_0
\end{aligned}, \tag{8}$$

where $\lambda_i = \Pr(C_{j_H}^1, \cdots, C_{j_H}^{n_{j_H}}, H_{j_H}^1, \cdots, H_{j_H}^i)$ and $i = 0, 1, \cdots, m_{j_H}$. If $i = 0$, $\lambda_0 = \Pr(C_{j_H}^1, \cdots, C_{j_H}^{n_{j_H}})$.

However, with the increasing number of child nodes in the $j_H$ th block, all nodes' state combinations will increase exponentially. Therefore, the joint probability table is compressed by the **Multistate Compression Algorithm** of Part I [1]. It is noteworthy that the function $\Psi(C_1, C_2, \cdots, C_n)$ as shown in figue 2 of Part I [1] should be changed to be the function $\Psi\{C_{j_H}^{c_{j_H}}, \Pr(C_{j_H}^{c_{j_H}} | \pi(C_{j_H}^{c_{j_H}}))\}$ in compressing the joint probability table. Suppose that the compression results are the compressed joint probability table $c\lambda_{m_{j_H}}$, the run accompanying dictionary $d_{r_{m_{j_H}}}$, the phrase accompany dictionary $d_{p_{m_{j_H}}}$, the set $RP^{m_{j_H}}$, and the set $S_{m_{j_H}}^{all}$. Based on $c\lambda_{m_{j_H}}$, the joint probability $\Pr_{j_H}$ is calculated by eliminating the root nodes $\{H_{j_H}^1, H_{j_H}^2, \cdots, H_{j_H}^{m_{j_H}}\}$ one by one. After eliminating the root node $H_{j_H}^{i+1}$ ($i = 0, \cdots, (m_{j_H} - 1)$), the intermediate factors $c\lambda_i$, $d_{r_i}$ and $d_{p_i}$ are constructed accordingly. The $j$ th row of $c\lambda_i$, denoted as $c\lambda_i^j$, is $\{run, q_i^j, n_{r_i}^j\}$ or $\{phrase, p_i^j, n_{p_i}^j\}$. Besides, the $q_i^j$ th row of $d_{r_i}$, denoted as $d_{r_i}^j$, is $\{q_i^j, r_i^j, L_{r_i}^j\}$ and the $p_i^j$ th row of $d_{p_i}$, denoted as $d_{p_i}^j$, is $\{p_i^j, v_{1_i}^j, v_{2_i}^j, L_{p_i}^j\}$. Suppose that $q_i^{\max 2}$ is the max row index in the run accompanying dictionary $d_{r_i}$. According to the $j$ th row of $c\lambda_{i+1}$, $c\lambda_i^j$ is constructed by Appendix C and Appendix D. The corresponding $d_{r_i}^j$ and $d_{p_i}^j$ is constructed by Table 1 and Table 2, respectively.

Table 1 Rules for constructing $d_{r_i}^j$

| switch | conditions | | | $r_i^j$ | $L_{r_i}^j$ |
|---|---|---|---|---|---|
| run | $H = 0$ | $F_r = N_i$ | | $r_i^j = R_{I-1}^{all} + r_{i+1}^j$ | $L_{r_i}^j = 1$ |
| | | $F_r > N_i$ | $V_r < N_i$ | $r_i^j = R_{I-1}^{all} + r_{i+1}^j$ | $L_{r_i}^j = 1$ |
| | | | $V_r \geq N_i$ | $r_i^j = R_{I-1}^{all} + r_{i+1}^j$ (also $r_i^{j+1} = N_i \times r_{i+1}^j$) | $L_{r_i}^j = 1$ (also $L_{r_i}^{j+1} = \frac{V_r - rem(V_r, N_i)}{N_i}$) |
| | $H = 1$ | $F_r < N_i$ | | -3 | $L_{r_i}^j = 0$ |
| | | $F_r = N_i$ | | $r_i^j = N_i \times r_{i+1}^j$ | $L_{r_i}^j = 1$ |
| | | $F_r > N_i$ | | $r_i^j = N_i \times r_{i+1}^j$ | $L_{r_i}^j = \frac{L_{r_{i+1}}^j - T_r}{N_i}$ |
| | $H > 1$ | $F_r < N_i$ | | -3 | $L_{r_i}^j = 0$ |
| | | $F_r = N_i$ | | $r_i^j = R_{I-1}^{all} + (N_i - H + 1) \times r_{i+1}^j$ | $L_{r_i}^j = 1$ |
| | | $F_r > N_i$ | $V_r < N_i$ | $r_i^j = R_{I-1}^{all} + (N_i - H + 1) \times r_{i+1}^j$ | $L_{r_i}^j = 1$ |
| | | | $V_r \geq N_i$ | $r_i^j = R_{I-1}^{all} + (N_i - H + 1) \times r_{i+1}^j$ (also $r_i^{j+1} = N_i \times r_{i+1}^j$) | $L_{r_i}^j = 1$ (also $L_{r_i}^{j+1} = \frac{V_r - rem(V_r, N_i)}{N_i}$) |

In Appendix C and Table 1, if the $j$ th row of $c\lambda_{i+1}$ is a run and $H = 0$, $F_r > N_i$ and $V_r \geq N_i$, both $c\lambda_i^j$ and $c\lambda_i^{j+1}$ are constructed at the same time. Because two rows of $c\lambda_i$ are constructed by one row of $c\lambda_{i+1}$, it is noteworthy that $c\lambda_i^{j+2}$ should be constructed by the $(j+1)$ th row of $c\lambda_{i+1}$. Besides, for this special situation, both $RP_{i+1}$ and $S_{i+1}^{all}$ should be updated synchronously. The detailed updating process is the same as the process in section 3.2 of Part I [1] (The final results are shown in equation (13) and (14) of Part I).

Table 2 Rules for constructing $d_{p_i}^{j}$

| switch | conditions | | | $v_{1_i}^{j}$ | $v_{2_i}^{j}$ | $L_{p_i}^{j}$ |
|---|---|---|---|---|---|---|
| phrase | $H=0$ | $V_p < N_i$ | | $R_{I-1}^{all} + v_{1_{i+1}}^{j}$ | -3 | 1 |
| | | $V_p \geq N_i$ | | $R_{I-1}^{all} + v_{1_{i+1}}^{j}$ | $N_i \times v_{2_{i+1}}^{j}$ | $\frac{V_p - rem(V_p, N_i)}{N_i} + 1$ |
| | $H=1$ | $F_p < N_i$ | | -3 | -3 | 0 |
| | | $F_p = N_i$ | | $v_{1_{i+1}}^{j} + (N_i - 1) \times v_{2_{i+1}}^{j}$ | -3 | 1 |
| | | $F_p > N_i$ | $V_p < 2N_i$ | $v_{1_{i+1}}^{j} + (N_i - 1) \times v_{2_{i+1}}^{j}$ | -3 | 1 |
| | | | $V_p \geq 2N_i$ | $v_{1_{i+1}}^{j} + (N_i - 1) \times v_{2_{i+1}}^{j}$ | $N_i \times v_{2_{i+1}}^{j}$ | $\frac{L_{p_{i+1}}^{j} - T_p}{N_i}$ |
| | $H>1$ | $F_p < N_i$ | | -3 | -3 | 0 |
| | | $F_p = N_i$ | | $R_{I-1}^{all} + v_{1_{i+1}}^{j} + (N_i - H) \times v_{2_{i+1}}^{j}$ | -3 | 1 |
| | | $F_p > N_i$ | $V_p < N_i$ | $R_{I-1}^{all} + v_{1_{i+1}}^{j} + (N_i - H) \times v_{2_{i+1}}^{j}$ | -3 | 1 |
| | | | $V_p \geq N_i$ | $R_{I-1}^{all} + v_{1_{i+1}}^{j} + (N_i - H) \times v_{2_{i+1}}^{j}$ | $N_i \times v_{2_{i+1}}^{j}$ | $\frac{V_p - rem(V_p, N_i)}{N_i} + 1$ |

*3.3 Multistate joint probability inference algorithm*

By eliminating root nodes of the $j_H$ th block one by one, the multistate joint probability inference algorithm is proposed for calculating the joint probability of all child nodes $\{C_{j_H}^{1}, C_{j_H}^{2}, \cdots, C_{j_H}^{n_{j_H}}\}$. The detailed calculation process is as follows.

The joint probability $\Pr(C_{j_H}^{1}, \cdots, C_{j_H}^{n_{j_H}}, H_{j_H}^{1}, \cdots, H_{j_H}^{m_{j_H}})$ is compressed by the multistate compression algorithm to get $c\lambda_{m_{j_H}}$, $d_{r_{m_{j_H}}}$, $d_{p_{m_{j_H}}}$, $RP_{m_{j_H}}$ and $S_{m_{j_H}}^{all}$. Based on $c\lambda_{m_{j_H}}$, the root nodes are eliminated one by one from $H_{j_H}^{m_{j_H}}$ to $H_{j_H}^{1}$. After eliminating the root node $H_{j_H}^{i+1}$, the new compressed intermediate factor $c\lambda_{i}^{new}$, the run accompanying dictionary $d_{r_i}^{new}$ and the phrase accompanying dictionary $d_{p_i}^{new}$ are created based on Appendix C, Appendix D, Table 1, and Table 2. After eliminating all root nodes, the following variables $c\lambda_i$, $d_{r_0}$, $d_{p_0}$, $RP_0$ and $S_0^{all}$ can be obtained. Finally, the joint probability $\Pr(C_{j_H}^{1}, \cdots, C_{j_H}^{n_{j_H}})$ can be gotten by decompressing $c\lambda_{N_Q+1}^{j_s}$ based on $d_{r_0}$, $d_{p_0}$ $RP_0$ and $S_0^{all}$. In summary, the multistate joint probability inference algorithm is shown in Table 3.

Table 3 Multistate joint probability inference algorithm

**Input** (1) Probability distributions of all multistate root nodes (2) $\Psi\{C_{j_H}^{c_{j_H}}, \Pr(C_{j_H}^{c_{j_H}} | \pi(C_{j_H}^{c_{j_H}}))\}$

(3) Relationship between the node $C_{j_H}^{c_{j_H}}$ and its parent nodes $\pi(C_{j_H}^{c_{j_H}})$

**Output** $\Pr(C_{j_H}^1, \cdots, C_{j_H}^{n_{j_H}})$

---

**Step 1** Compress the joint probability $\Pr(C_{j_H}^1, \cdots, C_{j_H}^{n_{j_H}}, H_{j_H}^1, \cdots, H_{j_H}^{m_{j_H}})$ by the **Multistate Compression Algorithm** of Part I [1] to get $c\lambda_{m_{j_H}}$, $d_{r_{m_{j_H}}}$, $d_{p_{m_{j_H}}}$, $RP_{m_{j_H}}$ and $S_{m_{j_H}}^{all}$.

**Step 2** For $i \leftarrow (m_{j_H} - 1)$ to 0, perform the calculation of (a)-(d) as shown in Appendix A to get $c\lambda_0$, $d_{r_0}$, $d_{p_0}$, $RP_0$ and $S_0^{all}$.

**Step 3** Decompress $c\lambda_0$ to get $\Pr(C_{j_H}^1, \cdots, C_{j_H}^{n_{j_H}})$ based on $d_{r_0}$, $d_{p_0}$, $RP_0$ and $S_0^{all}$.

---

*3.4 Proposed dependent multistate inference algorithm*

Based on VE, the dependent multistate inference algorithm is proposed to perform the inference of BN of the multistate system with dependent multistate components, as shown in Fig. 2. According to the block's definition, the BN in Fig. 2 can be equivalent to the BN in Fig. 3. Given the query node set $Q$ and the evidence set $E$ of multistate parent nodes, the conditional probability $\Pr(S | Q, E)$ is calculated by eliminating parent nodes of multistate node $S$ in Fig. 3. The detailed inference process is as follows.

Based on the node evidence set $E$, the probability distributions of all multistate nodes are updated at first. Secondly, according to the block's definition, the blocks can be found in the BN with dependent multistate components. Then, the joint probability of all child nodes in each block is calculated by the **Multistate Joint Probability Inference Algorithm**. Each block is equivalent to a multistate node and its marginal probability distribution is the joint probability. Finally, it is judged that whether the query node set $Q$ is empty or not. If the query node set $Q$ is not empty, the next step needs to judge whether the query node set $Q$ includes the nodes in blocks or not. The detailed judgment results have three situations, as shown in the following.

**Situation 1**: The query node set $Q$ is empty, i.e. $Q = \varnothing$.

For this situation, the probability distribution $\Pr(S | E)$ of the system node $S$ is the final result that needs to be calculated by given the root node evidence set $E$. Therefore, $\Pr(S | E)$ is calculated by eliminating all parent nodes one by one.

**Situation 2**: The query node set $Q$ is not empty, i.e. $Q \neq \emptyset$ but $Q$ does not include the child nodes in all blocks.

For this situation, the order of parent nodes in the NPT of the system node $S$ needs to be adjusted at first, i.e., all the nodes in $Q$ are reordered to the extreme left of NPT. Then, the conditional probability $\Pr(S|Q,E)$ is calculated by eliminating all non-query nodes one by one.

**Situation 3**: The query node set $Q$ is not an empty set, i.e., $Q \neq \emptyset$ and $Q$ includes at least one child node of one block.

For this situation, suppose that there are $\omega$ ($\omega \geq 1$) blocks and at least one child node in each of them is included in $Q$. Denote that the set of child nodes in the $k_Q$ th ($k_Q = 1, 2, \cdots, \omega$) block is $C_{k_Q}^Q$, and the $i_{k_Q}$ th ($i_{k_Q} = 1, 2, \cdots, n_{k_Q}$) child node in $C_{k_Q}^Q$ is $C_{k_Q}^{i_{k_Q}}$, where $n_{k_Q}$ is the number of child nodes of $C_{k_Q}^Q$. Therefore,

$$C_{k_Q}^Q = \{C_{k_Q}^1, C_{k_Q}^2, \cdots, C_{k_Q}^{i_{k_Q}}, \cdots C_{k_Q}^{n_{k_Q}}\}. \tag{9}$$

Besides, $Q$ includes $\tau$ ($\tau \geq 0$) dependent parent nodes $\{C_1^{Q\bar{H}}, C_2^{Q\bar{H}}, \cdots, C_{k_{\bar{Q}}}^{Q\bar{H}}, \cdots, C_\tau^{Q\bar{H}}\}$ of system node $S$, where $k_{\bar{Q}} = 1, 2, \cdots, \tau$. Define that $Q'$ is an extended query node set, i.e.

$$Q' = \{C_1^{Q\bar{H}}, C_2^{Q\bar{H}}, \cdots, C_\tau^{Q\bar{H}}, C_1^Q, C_2^Q, \cdots, C_\omega^Q\}, \tag{10}$$

Where

$$\begin{cases} C_1^Q = \{C_1^1, C_1^2, \cdots, C_1^{n_1}\} \\ C_2^Q = \{C_2^1, C_2^2, \cdots, C_2^{n_2}\} \\ \vdots \\ C_\omega^Q = \{C_\omega^1, C_\omega^2, \cdots, C_\omega^{n_\omega}\} \end{cases}. \tag{11}$$

The detailed calculation process includes seven steps as shown in (1)-(6).

(1) The parent nodes order in the NPT of the system node $S$ needs to be adjusted at first, i.e., all the nodes in the extended query node set $Q'$ are reordered to the extreme left of NPT. Then, the conditional probability $\Pr(S|Q')$ is calculated by eliminating all the nodes that are not included in the extended query node set $Q'$, i.e.

$$\Pr(S|Q') = \Pr(S|C_1^{Q\bar{H}}, C_2^{Q\bar{H}}, \cdots, C_\tau^{Q\bar{H}}, C_1^Q, C_2^Q, \cdots, C_\omega^Q) \tag{12}$$

(2) The joint probability $\Pr(S, Q')$ is calculated by equation (13), i.e.

$$\Pr(S, Q') = \Pr(S | Q') \times \Pr(Q'), \tag{13}$$

where $\Pr(Q') = \Pr(C_1^{Q\bar{H}}, \cdots, C_\tau^{Q\bar{H}}, C_1^Q, \cdots, C_\omega^Q)$. According to **Property 1** and **Property 2**, any two nodes of the set $\{C_1^{Q\bar{H}}, \cdots, C_\tau^{Q\bar{H}}, C_1^Q, \cdots, C_\omega^Q\}$ are independent of each other, i.e.

$$\begin{aligned}\Pr(Q') &= \Pr(C_1^{Q\bar{H}}, \cdots, C_\tau^{Q\bar{H}}, C_1^Q, \cdots, C_\omega^Q) \\ &= \prod_{k_Q=1}^{\omega} \Pr(C_{k_Q}^Q) \times \prod_{k_{\bar{Q}}=1}^{\tau} \Pr(C_{k_{\bar{Q}}}^{Q\bar{H}})\end{aligned} \tag{14}$$

Thus, the equation (13) can be changed to be

$$\begin{aligned}\Pr(S, Q') &= \Pr(S | Q') \times \Pr(Q') \\ &= \Pr(S | Q') \times \prod_{k_Q=1}^{\omega} \Pr(C_{k_Q}^Q) \times \prod_{k_{\bar{Q}}=1}^{\tau} \Pr(C_{k_{\bar{Q}}}^{Q\bar{H}}) \end{aligned}. \tag{15}$$

(3) For the nodes in $\{S, Q'\}$, if one node is included in the extended query node set $Q'$ but not included in the query node set $Q$, it will be reordered to the extreme right of the joint probability table of nodes $\{S, Q'\}$. It is noteworthy that the corresponding values of $\Pr(S, Q')$ are also adjusted accordingly as the order of the nodes in $\{S, Q'\}$ is adjusted. Based on the adjusted joint probability table, the joint probability $\Pr(S, Q)$ can be calculated by eliminating the nodes that are not included in the query node set $Q$, i.e.

$$\Pr(S, Q) = \sum_{Q'-Q} \Pr(S, Q'). \tag{16}$$

(4) Suppose that the intersection of the set $C_{k_Q}^Q$ and the query node set $Q$ is $C_{k_Q}^{HQ}$, i.e.

$$C_{k_Q}^{HQ} = Q \cap C_{k_Q}^Q. \tag{17}$$

All the nodes in the set $C_{k_Q}^{HQ}$ are reordered to the extreme left of the joint probability table of all child nodes in the $k_Q$ th ($k_Q = 1, 2, \cdots, \omega$) block. It is noteworthy that the corresponding values of $\Pr(C_{k_Q}^Q)$ are also adjusted accordingly as the order of the nodes in the set $C_{k_Q}^Q$ is adjusted. As shown in equation (18), the joint probability $\Pr(C_{k_Q}^{HQ})$ is calculated by eliminating all the nodes included in the set $C_{k_Q}^Q$ but not included in the set $C_{k_Q}^{HQ}$.

$$\Pr(C_{k_Q}^{HQ}) = \Pr(Q \cap C_{k_Q}^Q) \tag{18}$$

(5) According to **Property 2**, the joint probability of all query nodes is calculated by

$$\Pr(Q) = \prod_{k_Q=1}^{\omega} \Pr(C_{k_Q}^{HQ}) \times \prod_{k_{\bar{Q}}=1}^{\tau} \Pr(C_{k_{\bar{Q}}}^{Q\bar{H}}) \tag{19}$$

(6) Based on equations (16) and (19), the conditional probability is calculated by

$$\Pr(S|Q) = \frac{\Pr(S,Q)}{\Pr(Q)}. \tag{20}$$

Because the probability distributions of all nodes are updated according to the node evidence set $E$ at the beginning of inference, the equation (20) is equivalent to

$$\Pr(S|Q,E) = \frac{\Pr(S,Q)}{\Pr(Q)}. \tag{21}$$

In summary, the dependent multistate inference algorithm is shown in Table 4.

Table 4 Dependent multistate inference algorithm

| | |
|---|---|
| **Input** | (1) Probability distributions of all multistate root nodes |
| | (2) Relationship between the multistate node $S$ and its multistate parent nodes |
| | (3) NPTs of the multistate child nodes in blocks (4) Query node set $Q$ (5) Evidence set $E$ |
| **Output** | $\Pr(S|Q,E)$ |
| **Step 1** | Update the probability distributions of all multistate nodes by $E$. |
| **Step 2** | For $j_H \leftarrow \alpha$ to 1, do the following (I)-(IV) |
| | (I) Calculate the joint probability $\Pr(C_{j_H}^1,\cdots,C_{j_H}^{n_{j_H}})$ by **Multistate Joint Probability Inference Algorithm**. |
| | (II) The $j_H$ th block is equivalent to a multistate node $C'_{j_H}$. |
| | (III) $\Pr(C'_{j_H}) = \Pr(C_{j_H}^1,\cdots,C_{j_H}^{n_{j_H}})$. |
| | (IV) Calculate $T_{j_H}^C$ by equation (4). |
| | End |
| **Step 3** | If $Q$ meets the **Situation 1** or **Situation 2**, perform **Step 4**. If $Q$ meets the **Situation 3**, perform **Step 5**. |
| **Step 4** | Given $\{\Pr(C_1),\cdots,\Pr(C_{n_{\bar{H}}}),\Pr(C'_1),\cdots,\Pr(C'_\alpha)\}$, the query node set $Q$, the new evidence set $E' = \varnothing$ and $\Psi(C_1,\cdots,C_{n_{\bar{H}}},C'_1,\cdots,C'_\alpha)$, calculate $\Pr(Ch|Q,E)$ by **Independent Multistate Inference Algorithm** of Part I [1]. |
| **Step 5** | Perform the calculation of (1)-(14) as shown in Appendix B. |

## 4. Case Study

In section 4.1, the dependent situation of a satellite antenna pitch axis subsystem (the independent situation is shown in the Part I [1]) is used to validate the proposed dependent multistate inference algorithm. In section 4.2, the proposed algorithms of both two parts are applied to the reliability analysis of a more complex satellite attitude control system.

*4.1 Case 1*

*4.1.1 Background of case 1*

As described in the case study of Part I, a satellite antenna pitch axis subsystem ($PAS$) is composed of two stepper motors ($SM$), a drive shaft ($DF$), and a harmonic reducer ($HR$). The detailed data and structure relationships between components can refer to the case study of Part I. Different from Part I, this Part II considers that $DF$ and $HR$ have a common cause failure factor ($C$) (as shown in Fig. 4) in the actual pitch axis subsystem. The probability common cause failure factor occurring is $\Pr(C=2)=0.108$. If the common cause failure factor does not happen, the failure probabilities of $DF$ and $HR$ remain unchanged, i.e., $\Pr(DF=1|C=1)=0.019$ and $\Pr(HR=1|C=1)=0.013$. However, if the common cause failure factor occurs, the failure probabilities of $DF$ and $HR$ are $\Pr(DF=1|C=2)=0.187$ and $\Pr(HR=1|C=2)=0.143$.

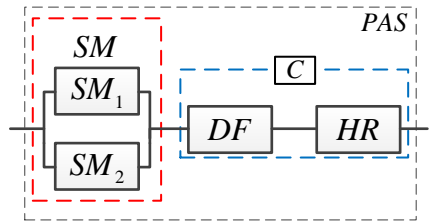

Fig. 4 Satellite antenna pitch axis subsystem with considering common cause failure

*4.1.2 Dependent situation inference*

According to Fig. 4, the BN of the pitch axis subsystem with considering common cause failure is constructed as shown in Fig. 5. According to the definition of the block, the node $C$, $DF$ and $HR$ in the red dotted box of Fig. 5 can be equivalent to be a block with four states (denoted as $DH$). The equivalent BN is shown in Fig. 6.

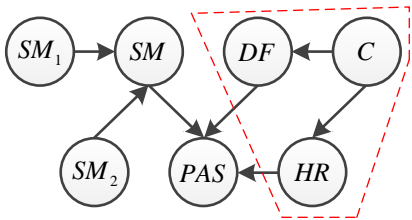 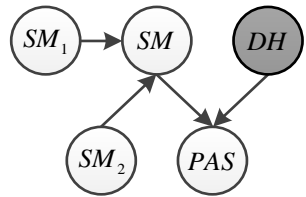

Fig. 5 BN of the pitch axis subsystem with considering common cause failure

Fig. 6 Equivalent BN of the pitch axis subsystem with considering common cause failure

According to the proposed multistate joint probability inference algorithm as shown in Table 3,

the joint probability of $DF$ and $HR$, i.e., the probability distribution of node $DH$, can be calculated as shown in Table 5. According to Table 5, the state combinations of the node $PAS$ in Fig. 6 are updated as shown in Table 7. By the proposed dependent multistate inference algorithm as shown in Table 4, the inference of BN in Fig. 6 is performed, and the probability distribution is shown in Table 6. To validate the correctness of the proposed dependent multistate inference algorithm, the BN of pitch axis subsystem is constructed by AgenaRisk software synchronously, and the results are shown in Fig. 7. Apparently, the results of the two methods are the same. Therefore, the correctness of the proposed dependent multistate inference algorithm is validated.

Table 5 Probability distribution of node $DH$

| $DH$ | | State | $\Pr(DH = \text{State})$ |
|---|---|---|---|
| $DF$ | $HR$ | | |
| 1 | 1 | 1 | 0.0239 |
| 1 | 2 | 2 | 0.1450 |
| 2 | 1 | 3 | 0.1051 |
| 2 | 2 | 4 | 0.7261 |

Table 6 Probability distribution of $PAS$

| State | $PAS = 1$ | $PAS = 2$ | $PAS = 3$ |
|---|---|---|---|
| $\Pr(PAS)$ | 0.0327 | 0.0582 | 0.9091 |

Table 7 NPT of the node $PAS$ in equivalent BN

| SM | DH | $\Pr(PAS \mid SM, DH)$ | | |
|---|---|---|---|---|
| | | $PAS = 1$ | $PAS = 2$ | $PAS = 3$ |
| 1 | 1 | 1 | 0 | 0 |
| 1 | 2 | 1 | 0 | 0 |
| 1 | 3 | 1 | 0 | 0 |
| 1 | 4 | 1 | 0 | 0 |
| 2 | 1 | 1 | 0 | 0 |
| 2 | 2 | 1 | 0 | 0 |
| 2 | 3 | 1 | 0 | 0 |
| 2 | 4 | 0 | 1 | 0 |
| 3 | 1 | 1 | 0 | 0 |
| 3 | 2 | 1 | 0 | 0 |
| 3 | 3 | 1 | 0 | 0 |
| 3 | 4 | 0 | 0 | 1 |

Besides, under different query node set $Q$, i.e. $\{SM = 3\}$, $\{DF = 2\}$, $\{HR = 2\}$, $\{SM = 3, DF = 2\}$, $\{SM = 3, HR = 2\}$ and $\{DF = 2, HR = 2\}$, the corresponding conditional probability of the pitch axis subsystem is calculated by the proposed algorithms. The results are shown in Fig. 8.

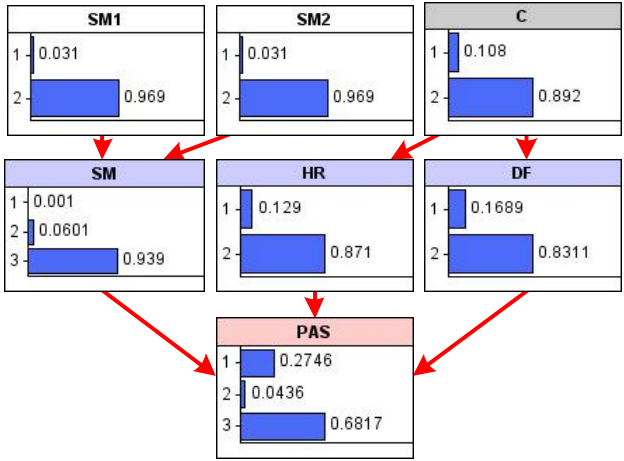 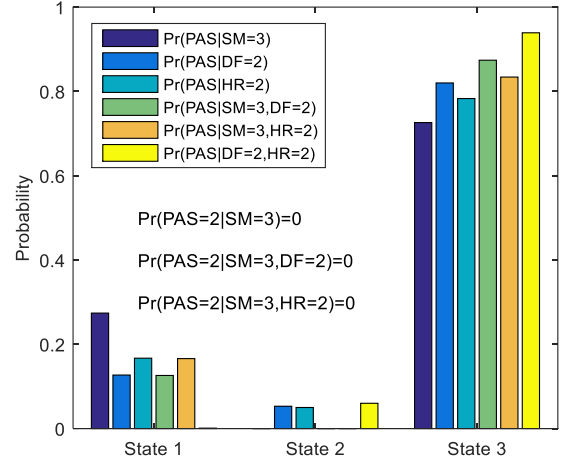

Fig. 7 Results of AgenaRisk software with considering common cause failure

Fig. 8 The conditional probabilities of the pitch axis subsystem's three states under different query node sets

In particular, when the query node set $Q$ is one of $\{SM=3\}$, $\{SM=3, DF=2\}$ and $\{SM=3, HR=2\}$, the corresponding conditional probability $\Pr(PAS=2|Q)$ is equal to 0 as shown in Fig. 8. This is led by the state definition of pitch axis subsystem. For example, the pitch axis subsystem's state 2 will not happen when it $SM$ is known to work normally (state 3). Therefore, the conditional probability $\Pr(PAS=2|SM=3)$ is equal to 0.

*4.2 Case 2*

*4.2.1 Background of case 2*

A satellite attitude control system includes a star sensor (2 single parts), leveling instrument (3 single parts), sun sensor (4 single parts), bearing frame (3 single parts), trestle1 (3 single parts), and trestle 2 (2 single parts) as shown in Fig. 9. All the state numbers and state definitions of the system and its components are shown in Fig. 9. The above six types of components can be divided into two parts: attitude sensor subsystem and structure subsystem. The attitude sensor subsystem is composed of star sensitive horizon equipment and sun sensor, where star sensitive horizon equipment includes star sensor and leveling instrument. The structure is composed of a bearing frame and two trestles.

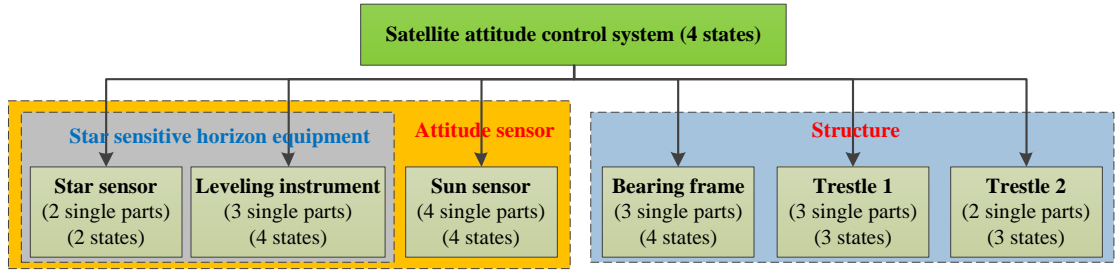

Fig. 9 Compositions of satellite attitude control system

For the single parts of six types of components, their compositions are shown in Fig. 10. The star sensor has four components (if two of the four components work, the star sensor works normally), the leveling instrument consists of two hot standby components in parallel, the sun sensor consists of two cold standby components in series, the trestle has two components in parallel, and the bearing frame includes three components which two parallel components are connected with one hot standby components in series. In addition to single star sensor and single trestle 1, all state definitions of system, subsystems and components are shown in Appendix E.

In Fig. 10, all components have two states, i.e., failure and normal. The failure probability of component changes as its working time changes. According to the component's electrical and mechanical properties, the lifetimes of the attitude control system's components obey exponential distribution and Weibull distribution as shown in equations (22) and (23), respectively. According to the history information of these components, the lifetime distributions of the components in Fig. 10 are shown in Table 8.

(a) Exponential distribution:

$$f(t) = \begin{cases} \theta e^{-\theta t}, & t > 0 \\ 0, & \text{other} \end{cases}, \quad (22)$$

where $\theta$ means the frequency that an event occurred in a unit of time.

(b) Weibull distribution:

$$f(t) = \begin{cases} \dfrac{\beta}{\eta}\left(\dfrac{t}{\eta}\right)^{\beta-1} e^{-\left(\frac{t}{\eta}\right)^{\beta}}, & t > 0 \\ 0, & \text{other} \end{cases}, \quad (23)$$

where $\beta$ ($\beta > 0$) is the shape parameter and $\eta$ ($\eta > 0$) is the proportional parameter.

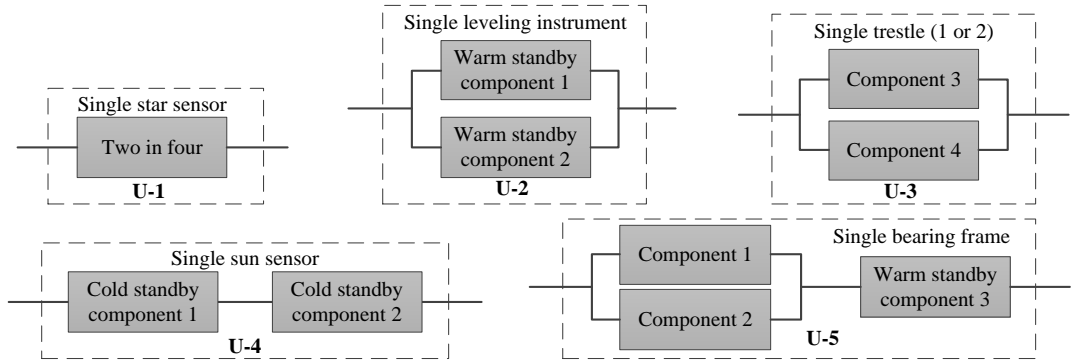

Fig. 10 Compositions of six single parts

Table 8 Components' lifetime distributions

| Component | Distribution | Parameter | Component | Distribution | Parameter |
|---|---|---|---|---|---|
| **A** | Exponential | $\theta = 1.36 \times 10^{-5}$ | **E** | Exponential | $\theta = 1.11 \times 10^{-5}$ |
| **B** | Exponential | $\theta = 1.03 \times 10^{-5}$ | **F** | Weibull | $\beta = 7.11$, $\eta = 8.78 \times 10^4$ |
| **C** | Weibull | $\beta = 6.02$, $\eta = 9.51 \times 10^4$ | **G** | Weibull | $\beta = 6.93$, $\eta = 8.39 \times 10^4$ |
| **D** | Weibull | $\beta = 5.03$, $\eta = 8.49 \times 10^4$ | **H** | Weibull | $\beta = 6.17$, $\eta = 8.62 \times 10^4$ |

**Remark**: **A**—Single star sensor's four components; **B**—Basic and optional components of hot standby component 1; **C**—Basic and optional components of hot standby component 2; **D**—Basic and optional components of cold standby component 1; **E**—Basic and optional components of cold standby component 2; **F**—Single bearing frame's two components; **G**—Basic and optional components of hot standby component 3; **H**—Two components of single trestle (1 or 2).

In actual satellite engineering, some components are dependent. In this case, the single star sensor and the single trestle 1 are dependent components. Considering the independent and dependent situation, the detailed state definitions of the single star sensor and the single trestle 1 are as follows:

**(1) Independent situation**

1) Single star sensor: As long as two of the four components work normally, the single star sensor can normally work (State 2). Otherwise, the single star sensor fails (State 1).

2) Single trestle 1: When two components of single trestle 1 work normally, the single trestle 1 works normally (State 3). When two components fail, the single trestle 1 fails (State 1). Otherwise, the working state of single trestle 1 downgrades one level (State 2).

**(2) Dependent situation**

For the dependent situation, there are three common cause failure factors, i.e. $H_1$, $H_2$ and $H_3$. $H_1$ affects two single star sensors' states simultaneously, and the states of three single trestles 1 are affected by $H_2$ and $H_3$. The happen probabilities of three factors are $\Pr(H_1 = 2) = 0.13$, $\Pr(H_2 = 2) = 0.21$ and $\Pr(H_3 = 2) = 0.16$, respectively. Besides, the state definitions of single star sensor and single trestle 1 are as follows:

1) Single star sensor: When the common cause failure factor $H_1$ happens, the single star sensor can normally work (State 2) under the condition that at least three components normally work. Otherwise, the single star sensor fails (State 1).

2) Single trestle 1: The single trestle 1 normally works (State 3) under the following two conditions: (a) all its components normally work; (b) both two common cause failure factors (i.e., $H_2$ and $H_3$) do not happens. The working state of single trestle 1 downgrades one level (State 2) under the following two conditions: (a) all its components normally work; (b) one of two common cause failure factors (i.e., $H_2$ and $H_3$) happens. Otherwise, the single trestle 1 fails (State 1).

*4.2.2 System BN reliability modeling*

As shown in Fig. 9, the satellite attitude control system has 17 components, and its components have $2^2 \times 4^3 \times 4^4 \times 4^3 \times 3^5 = 2^{22} \times 3^5$ state combinations. Therefore, the NPT size of the satellite attitude control system is $2^{22} \times 3^5 \times 4$. When the MATLAB 2016a software creates this NPT on a computer with 16GB RAM, the software will report an error. This is because the satellite attitude control system's NPT needs 30.4 GB (>16GB). Therefore, the BN model of the satellite attitude control system is modeled by the proposed multistate compression algorithm of Part I [1]. For the independent and dependent situations, the system BN reliability modeling is as follows:

(1) System BN reliability modeling for the independent situation

According to the satellite attitude control system's compositions, the BN reliability model of independent situation is constructed as shown in Fig. 11. The nodes numbered 1-58 are root nodes, and their marginal probability distributions are shown in Table 8. The nodes numbered 59-76 are non-root nodes, and their NPTs can be gotten by their state definitions.

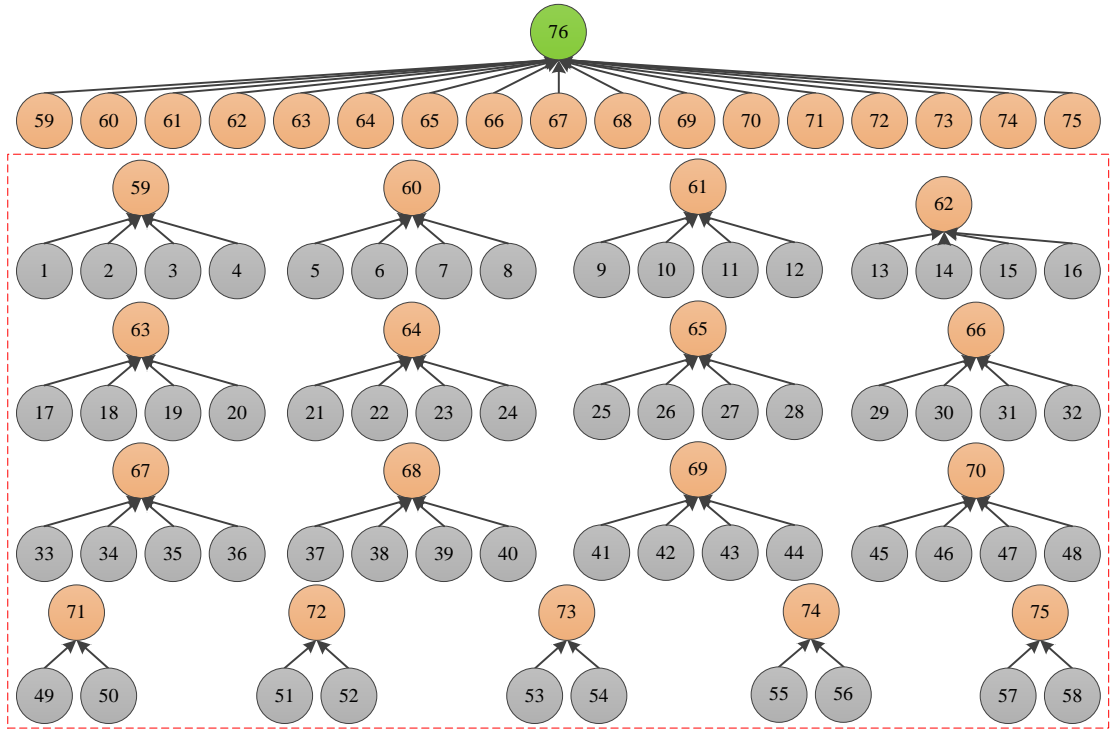

Remark: **1, 5**—Star sensor's first component; **2, 6**—Star sensor's second component; **3, 7**—Star sensor's third component; **4, 8**—Star sensor's fourth component; **9, 13, 17**—Basic components of hot standby component 1; **10, 14, 18**—Optional components of hot standby component 1; **11, 15, 19**—Basic components of hot standby component 2; **12, 16, 20**—Optional components of hot standby component 2; **21, 25, 29, 33**—Basic components of cold standby component 1; **22, 26, 30, 34**—Optional components of cold standby component 1; **23, 27, 31, 35**—Basic components of cold standby component 2; **24, 28, 32,36**—Optional components of cold standby component 2; **37, 41, 45**—Bearing frame's component 1; **38, 42, 46**—Bearing frame's component 2; **39, 43, 47**—Basic components of hot standby component 3; **40, 44, 48**—Optional components of hot standby component 3; **49, 51, 53**—Component 3 of Trestle 1; **50, 52, 54**—Component 4 of Trestle 1; **55, 57**—Component 3 of Trestle 2; **56, 58**—Component 4 of Trestle 2; **59, 60**—The first and second star sensor; **61, 62, 63**—The first, second and third leveling instrument; **64, 65, 66, 67**—The first, second, third and fourth sun sensor; **68, 69, 70**—The first, second and third bearing frame; **71, 72, 73**—The first, second and third trestle1; **74, 75**—The first and second trestle 2; **76**—Satellite attitude control system.

Fig. 11 BN reliability model of satellite attitude control system for the independent situation

(2) System BN reliability modeling for the dependent situation

According to the **Dependent situation,** as shown in section 4.2.1, the single star sensor (node 59 and 60) and the single trestle 1 (node 71, 72, and 73) are dependent components. Therefore, the relationships of these nodes in the system BN reliability model need to be adjusted accordingly, as shown in Fig. 12. According to the definition of block, two single star sensors and their parent nodes, three single trestle 1s and their parent nodes are equivalent to block 1 and block 2. In addition to the above five nodes, the relationships of other nodes shown in Fig. 11 remain unchanged.

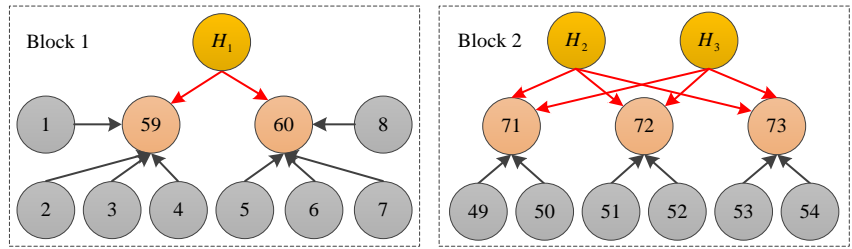

Fig. 12 The relationships of nodes with common failure factors for the dependent situation

(3) Discretization of the probability distribution

By integrating equation (22), the cumulative distribution function of the exponential distribution is shown in equation (24), where $F(t;\theta)$ is the component's failure probability when the component lifetime is $t$.

$$F(t;\theta) = \begin{cases} 1-e^{-\theta t} &, t > 0 \\ 0 &, \text{other} \end{cases} \quad (24)$$

By integrating equation (23), the cumulative distribution function of the Weibull distribution is shown in equation (25), where $F(t;\beta,\eta)$ is the component's failure probability when the component lifetime is $t$.

$$F(t;\beta,\eta) = \begin{cases} 1-e^{-\left(\frac{t}{\eta}\right)^{\beta}} &, t > 0 \\ 0 &, \text{other} \end{cases} \quad (25)$$

Select $\Delta t$ as the step size and calculate the probability distribution $P_{t_i}^{\tau}$ of the $\tau$ th component corresponding to $t_i$, where $\tau = 1, 2, \cdots, 58$, $t_i = t_0 + (i-1) \times \Delta t$ and $i = 1, 2, \cdots$. According to $\{P_{t_i}^{\tau} | \tau = 1, 2, \cdots, 58\}$, $\Delta t = 100$ and $t_0 = 0$, the inference of the BN in Fig. 11 is performed by the proposed algorithms.

*4.2.3 System reliability inference and analysis*

In this case, the BN reliability models of independent and dependent situations are inferred respectively by the proposed independent multistate inference algorithm of Part I [1] and the dependent multistate inference algorithm of this Part II. The results are discussed as follows:

(1) System probability distribution

The probability distributions of the satellite attitude control system for two situations are shown in Fig. 13 and Fig. 14, respectively. For both situations, the probability of normal work (i.e., state 4) decreases quickly as its lifetime increases. However, it is noteworthy that the probability of state 4 for the dependent situation decreases faster than the independent situation. For example, when the satellite attitude control system's lifetime arrives at $2 \times 10^4$ hours (about 2.28 years), the probability of state 4 for the independent and dependent situation are 0.6012 and 0.5325, respectively. Apparently, 0.6012 is more than 0.5325. Besides, the satellite attitude control system of the independent and dependent situations will completely fails when its lifetime arrives at $1.05 \times 10^5$ hours (about 11.99 years) and $1.0 \times 10^5$ hours (about 11.42 years), respectively. The reason for the above differences is that there are common cause failure factors between components. To further study the influence of common cause failure factors on system reliability, the following analysis is made.

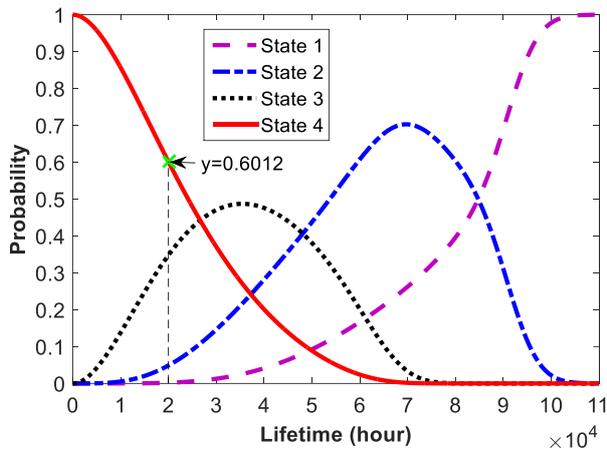 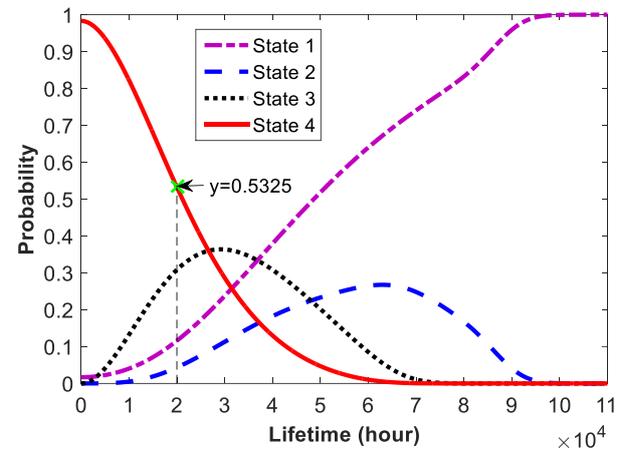

Fig. 13 Probability distribution of satellite attitude control system for independent situation

Fig. 14 Probability distribution of satellite attitude control system for dependent situation

(2) System reliability analysis

In satellite engineering, when the satellite attitude control system's working state downgrades one level, it can still regulate the satellite's attitude. Therefore, the sum of the satellite attitude control system's probabilities in state 3 and state 4 is used to assess the satellite attitude control system's reliability. The relationship between the satellite attitude control system's reliability and its lifetime is shown in Fig. 15.

Make the following assumption: To ensure that the satellite can operate normally on orbit, the satellite attitude control system's reliability must be no less than 0.5. Suppose the common cause failure factors are not considered in the reliability analysis. In that case, the engineers will mistakenly believe that the satellite attitude control system can meet the reliability requirement for $4.87 \times 10^4$ hours (about 5.56 years), as shown in Fig. 15. However, the common cause failure factors actually exist between some components. Therefore, when the satellite attitude control system's lifetime arrives at $4.87 \times 10^4$ hours, its actual reliability is 0.27 rather than 0.5. For the actual situation, i.e., the dependent situation, the satellite attitude control system can meet the reliability requirement for $3.7 \times 10^4$ hours (about 4.22 years), as shown in Fig. 15. From the above analysis, it can be found that the existence of common cause failure factors has some noticeable effect on the reliability of the satellite attitude control system.

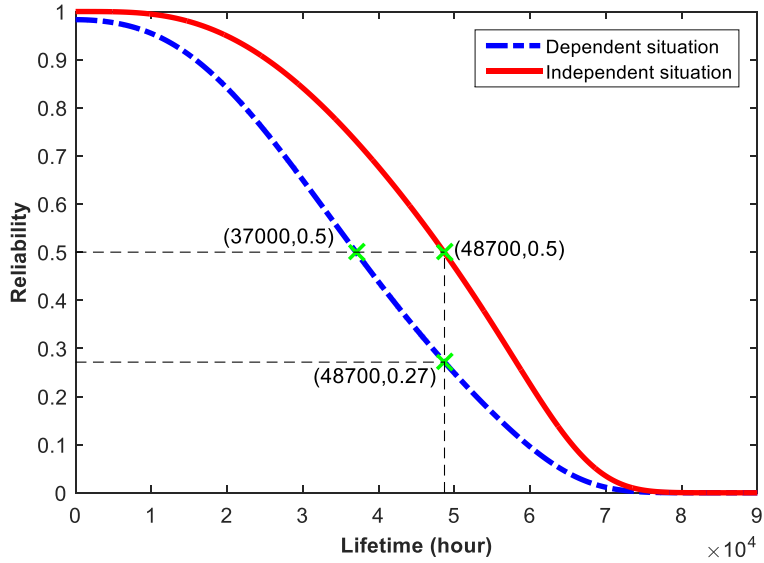

Fig. 15 The relationship between the satellite attitude control system's reliability and its lifetime

In summary, according to the above reliability inference and analysis, the proposed independent multistate inference algorithm of Part I [1] can perform the BN reliability inference of the complex multistate independent system. For the dependent situation, the proposed dependent multistate inference algorithm of this Part II also can do it. Besides, in analyzing the reliability of complex systems, over-simplifying the relationships between components (such as ignoring the common cause failure factors, etc.) will affect the analysis results.

## 5. Conclusions

As described in Part I [1], the multistate compression algorithm and the independent multistate algorithm are proposed for the BN reliability modeling and analysis of the complex multistate independent system. And this is also validated by the independent situation of case 2. In this Part II, a novel method is proposed for the complex multistate dependent system. For the dependent nodes and their parent nodes, they are equivalent to be a block. In processing the dependent nodes, the rules are proposed for constructing dependent intermediate factors, based on which the multistate joint probability inference algorithm is proposed to calculate the joint probability distribution of a block's all nodes. Then, based on the proposed multistate compression algorithm of Part I, the dependent multistate inference algorithm is proposed for the BN reliability modelling and analysis of the complex multistate dependent system. The use and accuracy of the proposed algorithms are demonstrated by the reliability analysis of the satellite antenna pitch axis subsystem. Finally, the proposed algorithms are applied to the reliability modeling and analysis of the satellite attitude control system in case 2. Besides, according to case 2, it can be known that the reliability analysis results of the complex system will be influenced by the over-simplifying the relationships between components (such as ignore the common cause failure factors, etc.). In summary, the dependent multistate inference algorithm's proposal makes the reliability modeling and analiysis of the complex multistate dependent system feasible. For future research, the authors will study how to simplify the calculation of dependent components further and improve the algorithms' computation efficiency.

**Acknowledgments**

This work was supported by the National Natural Science Foundation of China (No.51675525 and 11725211).

**References**

[1] Zheng X, Yao W, Xu Y, Chen X. Algorithms for Bayesian network modeling and reliability inference of complex multistate system: Part I –Independent system. Reliability Engineering & System Safety. 2020;202.
[2] Tien I, Kiureghian AD. Algorithms for Bayesian network modeling and reliability assessment of infrastructure systems. Reliability Engineering & System Safety. 2016;156:134-47.
[3] Zheng X, Yao W, Xu Y, Chen X. Improved compression inference algorithm for reliability analysis of complex multistate satellite system based on multilevel Bayesian network. Reliability Engineering & System Safety. 2019;189:123-42.
[4] O'Connor A, Mosleh A. A general cause based methodology for analysis of common cause and dependent failures in system risk and reliability assessments. Reliability Engineering & System Safety. 2016;145:341-50.
[5] Mi J, Li Y, Huang HZ, Liu Y, Zhang X. Reliability analysis of multi-state systems with common cause failure based on Bayesian Networks. Eksploatacja i Niezawodność – Maintenance and Reliability. 2013;15(2):169-75.
[6] Mi J, Li Y, Peng W, Huang H. Reliability analysis of complex multi-state system with common cause failure based on evidential networks. Reliability Engineering & System Safety. 2018;174:71-81.
[7] Mi J, Cheng Y, Song Y, Bai L, Chen K. Application of dynamic evidential networks in reliability analysis of complex systems with epistemic uncertainty and multiple life distributions. Annals of Operations Research. 2019.
[8] Mi J, Li YF, Yang YJ, Peng W, Huang HZ. Reliability assessment of complex electromechanical systems under epistemic uncertainty. Reliability Engineering & System Safety. 2016;152:1-15.
[9] Wojdowski L, Anders GJ. Substation reliability evaluation with dependent outages using Bayesian networks. 2016 International Conference on Probabilistic Methods Applied to Power Systems2016. p. 1-6.
[10] Pan Y, Ou S, Zhang L, Zhang W, Wu X, Li H. Modeling risks in dependent systems: A Copula-Bayesian approach. Reliability Engineering & System Safety. 2019;188:416-31.
[11] Tong Y, Tien I. Algorithms for Bayesian Network Modeling, Inference, and Reliability Assessment for Multistate Flow Networks. Journal of Computing in Civil Engineering. 2017;31:04017051.

**Appendix**

Appendix A The detailed calculation process (a)-(d)

(a) Calculate the row number $m_{i+1}$ of $c\lambda_{i+1}$.

(b) Eliminate $H_{j_H}^{i+1}$.

(c) For $j_u \leftarrow 1$ to $m_{i+1}$, do

① Construct $c\lambda_i^{new-j_u}$, $d_{r_i}^{new-j_u}$ and $d_{p_i}^{new-j_u}$ according to Appendix C, Appendix D, Table 1 and Table 2.

② Update $RP_{i+1}$ to get $RP_{i+1}^{update}$ and update $S_{i+1}^{all}$ to get $S_{i+1}^{update}$.

End, and output $c\lambda_i^{new}$, $d_{r_i}^{new}$, $d_{p_i}^{new}$, $RP_{i+1}^{update}$ and $S_{i+1}^{update}$.

(d) Based on $d_{r_i}^{new}$, $d_{p_i}^{new}$, $RP_{i+1}^{update}$ and $S_{i+1}^{update}$, decompress and compress $c\lambda_i^{new}$ row by row to get $c\lambda_i$, $d_{r_i}$, $d_{p_i}$, $RP_i$ and $S_i^{all}$.

Appendix B The detailed calculation process of **Step 5** in Table 4

(1) Construct the extended query node set $Q'$.

(2) Given $\{\Pr(C_1),\cdots,\Pr(C_{n_{\bar{B}}}),\Pr(C_1'),\cdots,\Pr(C_\alpha')\}$, the query node set $Q'$, the new evidence set $E' = \varnothing$ and $\Psi(C_1,\cdots,C_{n_{\bar{B}}},C_1',\cdots,C_\alpha')$, calculate $\Pr(S|Q')$ by the **Independent Multistate Inference Algorithm** of Part I [1].

(3) Calculate $\Pr(Q')$ and $\Pr(S,Q')$ by equation (14) and equation (15), respectively.

(4) Calculate the number $m_{\bar{Q}}$ of the nodes that are included in $Q'$ but not included in $Q$.

(5) Reorder the $m_{\bar{Q}}$ nodes to the extreme right of $\Pr(S,Q')$ to get $\Pr'(S,Q')$.

(6) Compress $\Pr'(S,Q')$ by the **Multistate Compression Algorithm** of Part I [1] to get $c\lambda_{m_{\bar{Q}}}$, $d_{r_{m_{\bar{Q}}}}$, $d_{p_{m_{\bar{Q}}}}$, $RP_{m_{\bar{Q}}}$ and $S_{m_{\bar{Q}}}^{all}$.

(7) For $i \leftarrow (m_{\bar{Q}}-1)$ to 0, do (a)-(d) as shown in Appendix A to get $c\lambda_0$, $d_{r_0}$, $d_{p_0}$, $RP_0$ and $S_0^{all}$. Then, decompress $c\lambda_0$ to get $\Pr(S,Q)$ based on $d_{r_0}$, $d_{p_0}$, $RP_0$ and $S_0^{all}$.

(8) For $k_Q \leftarrow 1$ to $\omega$, do the following (9)-(12).

(9) Calculate the set $C_{k_Q}^{HQ}$ by equation (18). Then, calculate the number of the nodes included in the set $C_{k_Q}^{HQ}$ but not included in the set $Q$, i.e. $l_{k_Q}$.

(10) Reorder the $l_{k_Q}$ nodes to the extreme right of $\Pr(C_{k_Q}^1,\cdots,C_{k_Q}^{n_{k_Q}})$ to get $\Pr'(C_{k_Q}^1,\cdots,C_{k_Q}^{n_{k_Q}})$.

(11) Compress $\Pr'(C_{k_Q}^1,\cdots,C_{k_Q}^{n_{k_Q}})$ by the **Multistate Compression Algorithm** of Part I [1] to get $c\lambda_{l_{k_Q}}$, $d_{r_{l_{k_Q}}}$, $d_{p_{l_{k_Q}}}$, $RP^{l_{k_Q}}$ and $S_{l_{k_Q}}^{all}$.

(12) For $i \leftarrow (l_{k_Q}-1)$ to 0, do (a)-(d) as shown in Appendix A to get $c\lambda_0$, $d_{r_0}$, $d_{p_0}$, $RP_0$ and $S_0^{all}$. Then, decompress $c\lambda_0$ to get $\Pr(C_{k_Q}^{HQ})$ based on $d_{r_0}$, $d_{p_0}$, $RP_0$ and $S_0^{all}$.

(13) Calculate $\Pr(Q)$ by equation (19).

(14) Calculate $\Pr(S|Q,E)$ by equation (21).

Appendix C Rules for constructing $c\lambda_i^j$ when the $j$ th row of $c\lambda_{i+1}^j$ is a run

| switch | condition | | | $J_{i+1}^j$ and $I$ | $q_{r_i}^j$ | $n_{r_i}^j$ | $R_{i+1}^j$ | $R^{all}$ |
|---|---|---|---|---|---|---|---|---|
| run | $H=0$ | $F_r = N_i$ | | $J_{i+1}^j = isme(RP_{i+1}^j, S_{i+1}^{all})$ $I = J_{i+1}^j(1)$ | $q_i^j = q_i^{\max 2}+1$ | $n_{r_{i+1}}^j$ | $0$ | $L_{RP} = length(RP_{i+1}^j)$ $\begin{cases} \text{for } i_{RP} = 1:L_{RP}, \text{do} \\ R_{J_{i+1}^j(1,i_{RP})}^{all} = R_{i+1}^j \\ \text{end} \end{cases}$ |
| | | $F_r > N_i$ | $V_r < N_i$ | | $q_i^j = q_i^{\max 2}+1$ | $n_{r_{i+1}}^j$ | $V_r \times r_{i+1}^j$ | |
| | | | $V_r \geq N_i$ | | $q_i^j = q_i^{\max 2}+1$ (also $q_i^{j+1} = q_i^{\max 2}+2$) | $n_{r_i}^j = n_{r_{i+1}}^j$ (also $n_{r_i}^{j+1} = n_{r_{i+1}}^j$) | If $rem(V_r, N_i) \neq 0$: $R_{i+1}^j = rem(V_r, N_i) \times r_{i+1}^j$ If $rem(V_r, N_i) = 0$: $R_{i+1}^j = 0$ | |
| | $H=1$ | $F_r < N_i$ | | | $q_i^j = q_i^{\max 2}+1$ | $n_{r_{i+1}}^j$ | $F_r \times r_{i+1}^j$ | |
| | | $F_r = N_i$ | | | $q_i^j = q_i^{\max 2}+1$ | $n_{r_{i+1}}^j$ | $0$ | |
| | | $F_r > N_i$ | | | $q_i^j = q_i^{\max 2}+1$ | $n_{r_{i+1}}^j$ | $T_r \times r_{i+1}^j$ | |
| | $H>1$ | $F_r < N_i$ | | | $q_i^j = q_i^{\max 2}+1$ | $n_{r_{i+1}}^j$ | $R_{I-1}^{all} + (F_r - H + 1) \times r_{i+1}^j$ | |
| | | $F_r = N_i$ | | | $q_i^j = q_i^{\max 2}+1$ | $n_{r_{i+1}}^j$ | $0$ | |
| | | $F_r > N_i$ | $V_r < N_i$ | | $q_i^j = q_i^{\max 2}+1$ | $n_{r_{i+1}}^j$ | $V_r \times r_{i+1}^j$ | |
| | | | $V_r \geq N_i$ | | $q_i^j = q_i^{\max 2}+1$ (also $q_i^{j+1} = q_i^{\max 2}+2$) | $n_{r_i}^j = n_{r_{i+1}}^j$ (also $n_{r_i}^{j+1} = n_{r_{i+1}}^j$) | If $rem(V_r, N_i) \neq 0$: $R_{i+1}^j = rem(V_r, N_i) \times r_{i+1}^j$ If $rem(V_r, N_i) = 0$: $R_{i+1}^j = 0$ | |

Appendix D Rules for constructing $c\lambda_i^j$ when the $j$ th row of $c\lambda_{i+1}^j$ is a phrase

| switch | condition | | | $J_{i+1}^j$ and $I$ | $p_i^j$ | $n_{p_i}^j$ | $R_{i+1}^j$ | $R^{all}$ |
|---|---|---|---|---|---|---|---|---|
| phrase | $H=0$ | $V_p < N_i$ | | $J_{i+1}^j = isme(RP_{i+1}^j, S_{i+1}^{all})$  $I = J_{i+1}^j(1)$ | $p_{i+1}^j$ | $n_{p_{i+1}}^j$ | $V_p \times v_{2_{i+1}}^j$ | $L_{RP} = length(RP_{i+1}^j)$  $\begin{cases} \text{for } i_{RP} = 1: L_{RP}, \text{do} \\ R^{all}_{J_{i+1}^j(1, i_{RP})} = R_{i+1}^j \\ \text{end} \end{cases}$ |
| | | $V_p \geq N_i$ | | | $p_{i+1}^j$ | $n_{p_{i+1}}^j$ | If $rem(V_p, N_i) \neq 0: R_{i+1}^j = rem(V_p, N_i) \times v_{2_{i+1}}^j$ | |
| | | | | | | | If $rem(V_p, N_i) = 0: R_{i+1}^j = 0$ | |
| | $H=1$ | $F_p < N_i$ | | | $p_{i+1}^j$ | $n_{p_{i+1}}^j$ | $v_{1_{i+1}}^j + (F_p - 1) \times v_{2_{i+1}}^j$ | |
| | | $F_p = N_i$ | | | $p_{i+1}^j$ | $n_{p_{i+1}}^j$ | 0 | |
| | | $F_p > N_i$ | $V_p < 2N_i$ | | $p_{i+1}^j$ | $n_{p_{i+1}}^j$ | $(V_p - N_i) \times v_{2_{i+1}}^j$ | |
| | | | $V_p \geq 2N_i$ | | $p_{i+1}^j$ | $n_{p_{i+1}}^j$ | If $rem(V_p, N_i) \neq 0: R_{i+1}^j = T_p \times v_{2_{i+1}}^j$ | |
| | | | | | | | If $rem(V_p, N_i) = 0: R_{i+1}^j = 0$ | |
| | $H>1$ | $F_p < N_i$ | | | $p_{i+1}^j$ | $n_{p_{i+1}}^j$ | $R_{I-1}^{all} + v_{1_{i+1}}^j + (F_p - H) \times v_{2_{i+1}}^j$ | |
| | | $F_p = N_i$ | | | $p_{i+1}^j$ | $n_{p_{i+1}}^j$ | 0 | |
| | | $F_p > N_i$ | $V_p < N_i$ | | $p_{i+1}^j$ | $n_{p_{i+1}}^j$ | $V_p \times v_{2_{i+1}}^j$ | |
| | | | $V_p \geq N_i$ | | $p_{i+1}^j$ | $n_{p_{i+1}}^j$ | If $rem(V_p, N_i) \neq 0: R_{i+1}^j = rem(V_p, N_i) \times v_{2_{i+1}}^j$ | |
| | | | | | | | If $rem(V_p, N_i) = 0: R_{i+1}^j = 0$ | |

# Appendix E Supplementary notes for Case 2

***State definition*:**

**(1) Star sensor**: When both single star sensors fail, the star sensor fails (State 1). Otherwise, the star sensor can normally work (State 2).

**(2) Single leveling instrument:** When all the basic and optional components of two hot standby components normally work, the single leveling instrument normally works (State 4). When only one of the basic components of two hot standby components fails, the single leveling instrument's working state downgrades one level (State 3). When all the basic and optional components of two hot standby components fail, the single leveling instrument fails (State 1). Otherwise, the single leveling instrument's working state downgrades two levels (State 2).

**(3) Leveling instrument:** When all single leveling instruments fail, the leveling instrument fails (State 1). When only one of three single leveling instruments normally works, the leveling instrument's working state downgrades one level (State 3). When two or three single leveling instruments normally work, the leveling instrument normally works (State 4). Otherwise, the leveling instrument's working state downgrades two levels (State 2).

**(4) Single sun sensor:** When all the basic components of two cold standby components normally work, the single sun sensor normally works (State 4). The single sun sensor working state downgrades one level (State 3) should meet two conditions synchronously: (a) One cold standby components' basic component fails and the corresponding optional component begins to work normally; (b) The other cold standby components' basic component can normally work. As long as both the basic and optional components of one cold standby component fail, the single sun sensor fails (State 1). Otherwise, the single sun sensor's working state downgrades two levels (State 2).

**(5) Sun sensor:** When all single sun sensors fail, the sun sensor fails (State 1). When one or two single sun sensors normally work, the sun sensor's working state downgrades one level (State 3). When three or four single sun sensors normally work, the sun sensor normally works (State 4). Otherwise, the sun sensor's working state downgrades two levels (State 2).

**(6) Single bearing frame:** When all components of the single bearing frame normally work, the single bearing frame normally works (State 4). The single bearing frame working state downgrades one level (State 3) should meet any one of the following two conditions: (a) Both component 1 and component 2 normally work, and one of the hot standby components' basic and the optional components normally works; (b) One of component 1 and component 2 normally work, and both the hot standby components' basic and the optional components can normally work. When both component 1 and component 2 fail or both the hot standby component's basic and optional components fail, the single bearing frame fails (State 1). Otherwise, the single bearing frame's working state downgrades two levels (State 2).

**(7) Bearing frame:** When all single bearing frames fail, the bearing frame fail (State 1). When only one of three single bearing frames normally works, the bearing frame's working state downgrades one level (State 3). When two or three single bearing frames normally work, the bearing frame normally works (State 4). Otherwise, the bearing frame's working state downgrades two levels (State 2).

**(8) Single trestle 2:** When two components of single trestle 2 normally work, the single trestle 2 normally works (State 3). When two components fail, the single trestle 2 fails (State 1). Otherwise, the working state of single trestle 2 downgrades one level (State 2).

**(9) Trestle 1**: When all single trestles fail, the trestle 1 fails (State 1). When two or three single trestles normally work, the trestle 1 normally works (State 3). Otherwise, the working state of trestle 1 downgrades one level (State 2).

**(10) Trestle 2:** When all single trestles fail, the trestle 2 fails (State 1). When at least one single trestle works normally, the trestle 2 normally works (State 3). Otherwise, the working state of trestle 2 downgrades one level (State 2).

**(11) Star sensitive horizon equipment:** When the star sensor and the leveling instrument work normally, the star sensitive horizon equipment normally works (State 3). As long as the star sensor or the leveling instrument fails, the star sensitive horizon equipment fails (State 1). Otherwise, the star sensitive horizon equipment's working state downgrades one level (State 2).

**(12) Attitude sensor:** When two components of the attitude sensor normally work, the attitude sensor normally works (State 4). The attitude sensor working state downgrades one level (State 3) should meet any one of the following two conditions: (a) The star sensitive horizon equipment normally work and the sun sensor's working state downgrades one level; (b) The star sensitive horizon equipment's working state downgrades one level and the sun sensor normally works. As long as the star sensitive horizon equipment or the sun sensor fails, the attitude sensor fails (State 1). Otherwise, the attitude sensor's working state downgrades two levels (State 2).

**(13) Structure:** Under the condition that the bearing frame normally works, when trestle 1 or trestle 2 normally works, the structure normally works (State 4). The structure working state downgrades one level (State 3) should meet any one of the following three conditions: (a) Both the bearing frame and trestle 1 normally work, and the working state of trestle 2 downgrades one level; (b) Both the bearing frame and trestle 2 normally work, and the working state of trestle 1 downgrades one level; (c) The bearing frame's working state downgrades one level, and both the trestle 1 and the trestle 2 normally work. Any two components of the structure fail, the structure fails (State 1). Otherwise, the structure's working state downgrades two levels (State 2).

**(14) Satellite attitude control system:** When both two subsystems work normally, the satellite attitude control system normally works (State 4). When one subsystem work normally and the other subsystem's working state downgrades one level, the satellite attitude control system working state downgrades one level (State 3). As long as the subsystem fails, the satellite attitude control system fails (State 1). Otherwise, the satellite attitude control system's working state downgrades two levels (State 2).